\documentclass[11pt]{article}

\usepackage[preprint]{acl}

\usepackage{times}
\usepackage{latexsym}
\usepackage{graphicx}
\usepackage{booktabs}
\usepackage{multirow}
\usepackage{subfigure}
\usepackage{algorithm}
\usepackage{algpseudocode}
\usepackage{amsmath,amssymb,amsthm}
\usepackage[inline]{enumitem}
\usepackage[T1]{fontenc}
\usepackage[utf8]{inputenc}

\usepackage{microtype}

\usepackage{inconsolata}

\usepackage{graphicx}
\usepackage[most]{tcolorbox}
\usepackage{xcolor}
\usepackage{booktabs}
\usepackage{amsmath}
\usepackage{pifont}
\usepackage{placeins}
\usepackage{dsfont}

\newcommand{\cmark}{\textcolor{cGreen}{\ding{51}}}
\newcommand{\xmark}{\textcolor{cRed}{\ding{55}}}

\newcommand{\polG}{\pi_{\theta}^{\mathsf{G}}}   % generator / reasoning policy
\newcommand{\polC}{\pi_{\theta}^{\mathsf{C}}}   % self-critic policy
\newcommand{\polR}{\pi_{\theta}^{\mathsf{R}}}   % self-corrector policy
\newcommand{\polTheta}{\pi_{\theta}}            % base policy
\DeclareMathOperator*{\Vote}{\mathcal{V}}        % plurality / majority vote
\DeclareMathOperator{\Extract}{\Phi}             % answer extraction map
\DeclareMathOperator{\Norm}{\eta}                % answer normalisation
\DeclareMathOperator*{\argmax}{arg\,max}
\newcommand{\Expect}{\mathbb{E}}
\newcommand{\Ind}{\mathds{1}}                    % indicator
\newcommand{\Qset}{\mathcal{Q}}
\newcommand{\Sset}{\mathcal{S}}
\newcommand{\Aset}{\mathcal{A}}
\newcommand{\simthr}{\tau_{\mathrm{sim}}}        % cosine-similarity clustering threshold
\newcommand{\embdim}{d_{\mathrm{emb}}}            % sentence-embedding dimension
\newcommand{\Tsample}{\tau}                       % rollout sampling temperature (alias)
\newcommand{\tin}{T_{\mathrm{in}}}
\newcommand{\tout}{T_{\mathrm{out}}}
\newcommand{\Pgen}{P_{\mathrm{gen}}}
\newcommand{\Prm}{P_{\mathrm{rm}}}
\newcommand{\Compute}{\mathcal{C}}
\newcommand{\CNG}{\eta}
 
\definecolor{cGreen}{RGB}{0,130,65}
\definecolor{cRed}{RGB}{175,50,50}
 
\definecolor{reasonBack}{RGB}{232,242,254}
\definecolor{reasonFrame}{RGB}{68,114,196}
 
\definecolor{critiqueBack}{RGB}{255,248,224}
\definecolor{critiqueFrame}{RGB}{185,135,25}
 
\definecolor{correctionBack}{RGB}{228,248,240}
\definecolor{correctionFrame}{RGB}{28,120,95}
 
\definecolor{problemBack}{RGB}{240,238,254}
\definecolor{problemFrame}{RGB}{95,85,195}
 
\definecolor{summaryBack}{RGB}{245,247,250}
\definecolor{summaryFrame}{RGB}{155,160,168}
 
\definecolor{boxRedBack}{RGB}{255,240,240}
\definecolor{boxRedFrame}{RGB}{180,80,80}
 
\newtcolorbox{reasonbox}[2][]{
  enhanced, breakable,
  colback=reasonBack, colframe=reasonFrame,
  coltitle=reasonFrame!70!black, fonttitle=\bfseries\small,
  title={#2}, boxrule=0.6pt, arc=1.5mm,
  left=6pt, right=6pt, top=5pt, bottom=5pt,
  before skip=4pt, after skip=4pt, #1
}
 
\newtcolorbox{critiquebox}[2][]{
  enhanced, breakable,
  colback=critiqueBack, colframe=critiqueFrame,
  coltitle=critiqueFrame!70!black, fonttitle=\bfseries\small,
  title={#2}, boxrule=0.6pt, arc=1.5mm,
  left=6pt, right=6pt, top=5pt, bottom=5pt,
  before skip=4pt, after skip=4pt, #1
}
 
\newtcolorbox{correctionbox}[2][]{
  enhanced, breakable,
  colback=correctionBack, colframe=correctionFrame,
  coltitle=correctionFrame!70!black, fonttitle=\bfseries\small,
  title={#2}, boxrule=0.6pt, arc=1.5mm,
  left=6pt, right=6pt, top=5pt, bottom=5pt,
  before skip=4pt, after skip=4pt, #1
}
 
\newtcolorbox{problembox}[2][]{
  enhanced, breakable,
  colback=problemBack, colframe=problemFrame,
  coltitle=problemFrame!70!black, fonttitle=\bfseries\small,
  title={#2}, boxrule=0.7pt, arc=2mm,
  left=7pt, right=7pt, top=6pt, bottom=6pt,
  before skip=6pt, after skip=6pt, #1
}
 
\newtcolorbox{summarybox}[2][]{
  enhanced, breakable,
  colback=summaryBack, colframe=summaryFrame,
  coltitle=black, fonttitle=\bfseries\small,
  title={#2}, boxrule=0.6pt, arc=2mm,
  left=7pt, right=7pt, top=6pt, bottom=6pt,
  before skip=6pt, after skip=6pt, #1
}
 
\newtcolorbox{warningbox}[2][]{
  enhanced, breakable,
  colback=boxRedBack, colframe=boxRedFrame,
  coltitle=black, fonttitle=\bfseries\small,
  title={#2}, boxrule=0.7pt, arc=2mm,
  left=7pt, right=7pt, top=6pt, bottom=6pt,
  before skip=6pt, after skip=6pt, #1
}

\renewtcolorbox{problembox}[1]{%
  enhanced, breakable,
  colback=problemBg, colframe=problemFrame, coltext=boxText,
  colbacktitle=problemFrame,
  coltitle=white,
  fonttitle=\bfseries\small\color{white},
  title={\color{white}#1},
  left=2mm, right=2mm, top=1.5mm, bottom=1.5mm,
  boxrule=0.6pt, arc=1mm
}
 
\renewtcolorbox{reasonbox}[1]{%
  enhanced, breakable,
  colback=reasonBg, colframe=reasonFrame, coltext=boxText,
  colbacktitle=reasonFrame,
  coltitle=white,
  fonttitle=\bfseries\small\color{white},
  title={\color{white}#1},
  left=2mm, right=2mm, top=1.5mm, bottom=1.5mm,
  boxrule=0.6pt, arc=1mm
}
 
\renewtcolorbox{critiquebox}[1]{%
  enhanced, breakable,
  colback=critiqueBg, colframe=critiqueFrame, coltext=boxText,
  colbacktitle=critiqueFrame,
  coltitle=white,
  fonttitle=\bfseries\small\color{white},
  title={\color{white}#1},
  left=2mm, right=2mm, top=1.5mm, bottom=1.5mm,
  boxrule=0.6pt, arc=1mm
}
 
\renewtcolorbox{correctionbox}[1]{%
  enhanced, breakable,
  colback=correctionBg, colframe=correctionFrame, coltext=boxText,
  colbacktitle=correctionFrame,
  coltitle=white,
  fonttitle=\bfseries\small\color{white},
  title={\color{white}#1},
  left=2mm, right=2mm, top=1.5mm, bottom=1.5mm,
  boxrule=0.6pt, arc=1mm
}
 
\renewtcolorbox{summarybox}[1]{%
  enhanced, breakable,
  colback=summaryBg, colframe=summaryFrame, coltext=boxText,
  colbacktitle=summaryFrame,
  coltitle=white,
  fonttitle=\bfseries\small\color{white},
  title={\color{white}#1},
  left=2mm, right=2mm, top=1.5mm, bottom=1.5mm,
  boxrule=0.6pt, arc=1mm
}

\providecolor{cGreen}{RGB}{0, 120, 60}
\providecolor{cRed}{RGB}{180, 30, 30}
\providecolor{problemFrame}{RGB}{60, 60, 60}
\providecolor{reasonFrame}{RGB}{30, 90, 160}
\providecolor{critiqueFrame}{RGB}{200, 110, 0}
\providecolor{correctionFrame}{RGB}{0, 120, 60}
\providecolor{summaryFrame}{RGB}{90, 50, 130}
\providecolor{problemBg}{RGB}{245, 245, 245}
\providecolor{reasonBg}{RGB}{235, 243, 252}
\providecolor{critiqueBg}{RGB}{253, 244, 230}
\providecolor{correctionBg}{RGB}{232, 246, 237}
\providecolor{summaryBg}{RGB}{243, 238, 250}
\providecolor{boxText}{RGB}{20, 20, 20}
 
\providecommand{\cmarkw}{\textcolor{white}{\ding{51}}}

\providecommand{\cmark}{\textcolor{cGreen}{\ding{51}}}
\providecommand{\xmark}{\textcolor{cRed}{\ding{55}}}

\title{Refining Over Resampling: Test-Time Self-Correction for LLM Reasoning}

\author{
  \textbf{Ahsan Bilal\textsuperscript{1}},
  \textbf{Muhammad Ahmed Mohsin\textsuperscript{2}},
  \textbf{Muhammad Umer\textsuperscript{2}},
  \textbf{Lena Trigg\textsuperscript{1}},
\\
  \textbf{Ali Subhan\textsuperscript{3}},
  \textbf{Muhammad Ali\textsuperscript{4}},
  \textbf{Dean F. Hougen\textsuperscript{1}}
\\
\\
  \textsuperscript{1}University of Oklahoma,
  \textsuperscript{2}Stanford University,
\\
  \textsuperscript{3}Universitat Pompeu Fabra,
  \textsuperscript{4}Air University
\\
  \small{
    \textbf{Correspondence:} \href{mailto:ahsan.bilal-1@ou.edu}{ahsan.bilal-1@ou.edu}
  }
}

\begin{document}
\maketitle
\begin{abstract}
Test-time scaling improves LLM reasoning by using additional inference compute, but wider sampling alone can suffer from diminishing returns: new rollouts often repeat existing answer patterns instead of adding useful reasoning diversity. Verifier-based selection offers an alternative, but its performance depends on the calibration of an external reward model. We propose a verifier-free breadth--depth refinement framework that uses test-time compute to both explore and improve candidate solutions. The method samples multiple independent reasoning rollouts, refines each rollout through iterative self-critique and self-correction, and aggregates the refined answers by majority voting. Breadth preserves diverse initial attempts, while depth repairs local reasoning errors before aggregation. Across AIME24, AIME25, AMC, OlympiadBench, and MATH500, our method consistently improves over greedy decoding, majority voting, verifier-based best-of-$N$, beam search, and lookahead decoding across multiple open-weight models. For instance, with Qwen2.5-1.5B, accuracy increases from the strongest verifier-based baseline to $58.0\%$ on MATH500, and from $25.0\%$ to $32.5\%$ on AMC. These results show that test-time compute can be more effective when used to refine sampled trajectories rather than only to sample more candidates or rely on verifier-guided selection.
\end{abstract}

\section{Introduction}
Test-time scaling has emerged as a practical way to improve large language model (LLM) reasoning without increasing model size or updating model parameters. Instead of relying on a single response, these methods allocate additional inference compute to generate, search, verify, or refine candidate reasoning trajectories before producing a final answer~\cite{beirami2024theoretical, zuo2026ttrl, inoue2026wider, wang2026nablareasoner}. This paradigm is especially important for mathematical reasoning, where early mistakes in a chain of reasoning can propagate to the final answer, and where additional inference compute can expose alternative solution paths that are not available from a single greedy generation.

\begin{figure}[t]
    \centering
    \begin{minipage}[t]{0.5\linewidth}
        \centering
        \includegraphics[width=\linewidth]{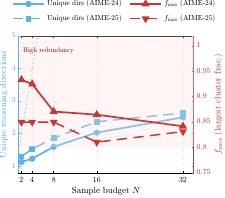}
        \caption*{(a) Dominant cluster mass.}
    \end{minipage}\hfill
    \begin{minipage}[t]{0.5\linewidth}
        \centering
        \includegraphics[width=\linewidth]{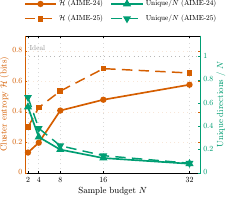}
        \caption*{(b) Diversity saturation.}
    \end{minipage}
    \caption{\textbf{Sampling redundancy under increasing budget.} As the sample budget grows, rollouts remain concentrated in a few semantic reasoning clusters, while the number of unique reasoning directions per sample decreases.}
    \label{fig:motivation}
\end{figure}

A common strategy is to sample multiple reasoning traces and aggregate them using majority voting or verifier-based selection. Self-consistency improves reliability by selecting the most frequent answer among sampled reasoning paths~\cite{wang2022self}, while verifier-based approaches use outcome or process-level reward models to score candidate solutions~\cite{cobbe2021training, lightman2023let, li2023making, wang2024math}. However, these approaches have two limitations. First, additional samples do not always provide genuinely new reasoning evidence; as shown in Figure~\ref{fig:motivation}, larger sample budgets can repeatedly produce variants of a small number of dominant reasoning directions. Second, verifier-based selection introduces dependence on reward models whose calibration errors can directly affect final-answer selection~\cite{dorner2025roc, li2025fixing, zhang2025lessons}.

Self-critique and self-refinement provide an alternative direction: rather than only sampling more candidates or relying on an external scorer, the model can use additional test-time compute to inspect and improve its own reasoning. Prior work has shown that iterative feedback and revision can improve LLM outputs~\cite{madaan2023self}, and later methods have explored self-correction, reflection, progressive refinement, and Monte Carlo refinement for reasoning tasks~\cite{shinn2024reflexion, zhang2024accessing, du2025think, yuan2025reinforce}. At the same time, intrinsic self-correction remains difficult: LLMs may fail to identify their own reasoning errors and can even degrade their answers when refinement is applied naively~\cite{huang2024large}. This suggests that effective refinement should not rely on a single correction attempt but should instead use test-time compute in a structured way.

Recent work improves self-correction by training models to revise or verify their own outputs: SCoRe uses multi-turn reinforcement learning to strengthen intrinsic self-correction~\cite{kumar2024training}, while ReVISE learns a stop-or-refine policy through self-verification~\cite{lee2025revise}; however, both require additional training, learned verification, or reward-driven optimization. This limits their use as simple, model-agnostic test-time methods. We instead propose a training-free and verifier-free breadth-depth refinement framework that uses only the base model at inference time. It samples multiple independent rollouts, refines each one through iterative self-critique and self-correction, and aggregates the refined answers by majority vote.

\paragraph{Contributions.}
Our contributions are threefold. First, we identify two limitations of width-only test-time scaling: diversity saturation, where larger sampling budgets revisit existing semantic reasoning clusters, and per-trace hallucination, where each i.i.d.\ rollout remains exposed to the same perturbation mechanism. Second, we propose a breadth-depth refinement framework that addresses both issues: breadth preserves diverse initial reasoning paths through $N$ independent rollouts, while depth applies $D$ rounds of self-critique and self-correction to repair errors before majority-vote aggregation. The method requires no external verifier, PRM, learned stopping policy, or additional training. Third, we evaluate the framework on five mathematical reasoning benchmarks and four open-weight models, showing consistent gains over greedy decoding, majority voting, verifier-based best-of-$N$, beam search, and lookahead baselines, together with compute-normalized analysis and refinement dynamics diagnostics.

\begin{figure*}[t]
    \centering
    \includegraphics[width=\linewidth]{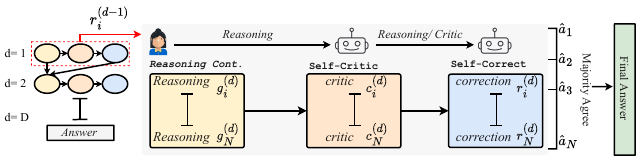}
\caption{\textbf{Overview of the proposed breadth-depth refinement framework.} Each of the $N$ rollouts is refined through $D$ depth layers, each consisting of three stages: (1)~\emph{reasoning continuation} ($g_i^{(d)}$) by $\polG$, (2)~\emph{self-critique} ($c_i^{(d)}$) by $\polC$, and (3)~\emph{self-correction} ($r_i^{(d)}$) by $\polR$. Terminal traces are aggregated by majority voting to produce the final answer $\hat{a}$. No verifier or additional training is required. }
    \label{fig:architecture}
\end{figure*}
\section{Problem Motivation}
\label{sec:motivation}
 
\paragraph{Diversity saturation under fixed-budget sampling.}
Let $\polTheta$ denote a language-model policy and let $\{r_i^{(0)}\}_{i=1}^{N} \stackrel{\text{i.i.d.}}{\sim} \polTheta(\cdot\mid x)$ be $N$ independently sampled reasoning rollouts for an input problem $x$. Repeated sampling followed by majority voting is a standard test-time scaling strategy~\cite{wang2022self,beirami2024theoretical,inoue2026wider}. Its benefit, however, depends on whether additional samples provide genuinely new reasoning evidence rather than variants of the same answer pattern.

We group initial rollouts into sentence-embedding-based clusters by semantic similarity. Let $\{\mathcal{C}_j\}_{j\ge 1}$ be the clusters, with empirical mass $p_j(N)=N^{-1}\sum_{i=1}^{N}\Ind[r_i^{(0)}\in\mathcal{C}_j]$. We track the number of realized clusters, $\mathcal{R}(N)$, and entropy, $\mathcal{H}(N)=-\sum_j p_j(N)\log_2 p_j(N)$. These measure the number of distinct reasoning directions and how evenly samples are distributed across them. Figure~\ref{fig:motivation} shows that diversity grows slowly as $N$ increases; additional details are provided in Appendix~\ref{app:metrics:redundancy}.
\begin{equation}
\setlength{\jot}{1pt}
\begin{alignedat}{2}
  \frac{\mathcal{R}(N)}{N}
  &\xrightarrow{N\to\infty}{} 0, \qquad&
  \frac{\mathrm{d}\mathcal{H}(N)}{\mathrm{d}N}
  &\xrightarrow{N\to\infty}{} 0 .
\end{alignedat}
\label{eq:saturation}
\end{equation}
For example, on AIME-24 the number of unique clusters increases only from roughly $1.2$ at $N{=}2$ to about $2.5$ at $N{=}32$, with a similar trend on AIME-25. Thus, larger-width budgets increasingly revisit existing semantic reasoning clusters instead of expanding the set of reasoning directions. This motivates using additional test-time compute not only to resample but also to refine the sampled trajectories.

\paragraph{Hallucination as a systematic accuracy floor.}
A complementary limitation of width-only sampling is that each sampled trace can still suffer from the same local generation errors. Consider a problem with $k$ possible answer classes. Following \citet{liu2024large}, let $\lambda \in (1/k,1]$ denote the model's latent accuracy, i.e., the probability that its intended answer is correct, and let $h\in(0,1)$ denote the probability that the realized answer is perturbed away from this intended answer to one of the remaining $k{-}1$ classes. For a single sampled answer $\hat{a}_{\mathrm{single}}\sim\polTheta(\cdot\mid x)$ and gold answer $a^{\star}$, the expected single-trace accuracy is
\begin{equation}
\begin{aligned}
  \Expect\!\bigl[\Ind[\hat{a}_{\mathrm{single}}=a^{\star}]\bigr]
  &= \lambda(1-h) + \frac{h(1-\lambda)}{k-1} \\
  &= \lambda\!\left(1 - \frac{hk}{k-1}\right) \\
  &\quad + \frac{h}{k-1} < \lambda .
\end{aligned}
\label{eq:hallucination}
\end{equation}
The inequality holds because, when $\lambda>1/k$, the intended answer is more likely to be correct than random. A perturbation, therefore, destroys correct intended answers more often than it accidentally fixes incorrect ones. As a result, the realized first-pass answer systematically underestimates the model's latent accuracy. Width-only scaling does not remove this effect because every i.i.d.\ rollout is exposed to the same perturbation mechanism. This motivates refining each sampled trajectory before aggregation, rather than only drawing more samples.

\paragraph{Structured refinement as the principled remedy.}
Eqs.~\eqref{eq:saturation} and~\eqref{eq:hallucination} identify two limitations of width-only scaling: redundancy in the sample space and per-trace hallucination bias. Both motivate adding structured depth. Prior self-refinement methods show that models can improve generated outputs by critiquing and revising their own responses~\cite{madaan2023self,shinn2024reflexion,zhang2024accessing}, but naive self-correction can also degrade reasoning quality~\cite{huang2024large}. \citet{liu2024large} show that second-pass correction can reduce the effective hallucination rate, shrinking the gap in Eq.~\eqref{eq:hallucination}. This motivates applying $D$ refinement depths to each rollout, where each depth consists of a self-critique followed by self-correction. Breadth addresses Eq.~\eqref{eq:saturation} by preserving diverse starting directions across $N$ independent rollouts, while depth addresses Eq.~\eqref{eq:hallucination} by refining each trajectory before aggregation. Majority voting over the refined outputs then avoids dependence on an auxiliary verifier or reward-model calibration~\cite{cobbe2021training,lightman2023let,dorner2025roc,zhang2025lessons}.

\begin{algorithm}[t]
\caption{Breadth-Depth Test-Time Refinement}
\label{alg:method}
\begin{algorithmic}[1]
\Require Problem $x$;\ policy $\polTheta$ with generator/critic/corrector roles
         $\polG,\polC,\polR$;\ rollouts $N$;\ depth $D$;\
         sampling temperature $\Tsample > 0$
\Ensure  Final prediction $\hat{a}$
 
\Statex \textcolor{blue}{\textbf{\# Initialization}}
\For{$i = 1$ \textbf{to} $N$}  \Comment{parallel\,/\,batched}
  \State $r_i^{(0)} \sim \polG(\cdot \mid x)$
         \hfill\Comment{i.i.d.\ sample at temperature $\Tsample$}
\EndFor
 
\Statex \textcolor{blue}{\textbf{\# Iterative refinement}}
\For{$d = 1$ \textbf{to} $D$}
  \For{$i = 1$ \textbf{to} $N$}  \Comment{all $N$ processed in one batched call}
    \State $g_i^{(d)} \sim \polG\!\bigl(\cdot \mid x,\, r_i^{(d-1)}\bigr)$
           \hfill\Comment{Eq.~\eqref{eq:reasoning}: reasoning continuation}
    \State $c_i^{(d)} \sim \polC\!\bigl(\cdot \mid x,\, g_i^{(d)}\bigr)$
           \hfill\Comment{Eq.~\eqref{eq:critique}: self-critique}
    \State $r_i^{(d)} \sim \polR\!\bigl(\cdot \mid x,\, g_i^{(d)},\, c_i^{(d)}\bigr)$
           \hfill\Comment{Eq.~\eqref{eq:correction}: self-correction}
  \EndFor
\EndFor
 
\Statex \textcolor{blue}{\textbf{\# Aggregation}}
\For{$i = 1$ \textbf{to} $N$}
  \State $\hat{a}_i \gets \Extract\!\bigl(r_i^{(D)}\bigr)$
         \hfill\Comment{answer extraction}
\EndFor
\State $\hat{a} \gets \Vote\!\bigl(\hat{a}_1,\ldots,\hat{a}_N\bigr)$
       \hfill\Comment{plurality vote; no verifier}
\State \Return $\hat{a}$
\end{algorithmic}
\end{algorithm}
\section{Methodology}
\label{sec:method}
 
\subsection{Overview} \label{sec:method:overview}
Given an input problem $x$, our framework allocates test-time compute along two complementary axes: \emph{breadth}, by sampling $N$ independent reasoning rollouts, and \emph{depth}, by refining each rollout for $D$ iterative steps. Let $\polTheta$ denote the underlying language-model policy with parameters $\theta$; the same model is used throughout, with role-conditioned prompts defining a generator $\polG$, a critic $\polC$, and a corrector $\polR$. Each rollout is first initialized by the generator and then updated through a repeated three-stage refinement cycle: the generator continues or rewrites the previous corrected trace, the critic identifies possible logical, arithmetic, or structural errors, and the corrector revises the trace conditioned on the generated reasoning and its critique. The resulting terminal traces are passed to a deterministic answer-extraction and plurality-voting step, described in Section~\ref{sec:method:agg}. Thus, the method uses breadth to preserve diverse solution attempts and depth to repair errors within each trajectory, while requiring no external verifier, process reward model, learned stopping policy, or additional training. Figure~\ref{fig:architecture} summarizes the overall breadth-depth refinement pipeline, and Algorithm~\ref{alg:method} gives the complete procedure.

\begin{table*}[t]
\centering
\small
\caption{\textbf{Accuracy (\%) on mathematical reasoning benchmarks.} $N{=}8$ for all sampling-based methods; RM@8 uses Qwen2.5-Math-RM-32B as the verifier.}
\label{tab:selected_math_results_transposed}
\begin{tabular}{llccccc}
\hline
\textbf{Model} & \textbf{Method} & \textbf{AIME24} & \textbf{AIME25} & \textbf{AMC} & \textbf{OlyBench} & \textbf{MATH} \\
\hline
\multirow{6}{*}{\textbf{Qwen2.5-Math-7B}}
& Greedy           & 3.3  & 13.33 & 55.0 & 35.0 & 75.0 \\
& Maj@8            & 10.0  & 6.67  & 60.0 & 39.5 & 79.0 \\
& RM@8             & \textbf{13.33} & 13.33 & 62.5 & 41.0 & 81.0 \\
& Beam ($B=8$)     & 10.0  & 10.0  & 62.5 & 38.5 & 77.0 \\
& Lookahead        & 10.0  & 6.67  & 60.0 & 39.0 & 78.3 \\
& \textbf{Ours}    & 10.0  & \textbf{16.67} & \textbf{67.5} & \textbf{47.2} & \textbf{81.6} \\
\hline
\multirow{6}{*}{\textbf{Qwen2.5-1.5B}}
& Greedy           & 0.0  & \textbf{6.67} & 25.0 & 7.5  & 26.2 \\
& Maj@8            & 3.33 & 3.33 & 17.5 & 12.5 & 29.4 \\
& RM@8             & 3.33 & 3.33 & 25.0 & 13.5 & 29.6 \\
& Beam ($B=8$)     & 3.33 & 3.33 & 20.5 & 13.0 & 29.0 \\
& Lookahead        & 3.33 & 3.33 & 20.5 & 14.0 & 28.8 \\
& \textbf{Ours}    & \textbf{6.67} &  \textbf{6.67} & \textbf{32.5} & \textbf{24.5} & \textbf{58.0} \\
\hline
\multirow{6}{*}{\textbf{Ministral-8B}}
& Greedy           & 3.33 & 0.0  & 20.0 & 20.0 & 56.8 \\
& Maj@8            & 0.0  & 0.0  & 27.5 & 23.5 & 59.2 \\
& RM@8             & \textbf{6.67} & 3.33 & 30.0 & 24.5 & 61.4 \\
& Beam ($B\!=\!8$) & 0.0  & \textbf{6.67} & 27.5 & 26.5 & 58.4 \\
& Lookahead        & 3.33 & 3.33 & 25.0 & 24.5 & 59.4 \\
& \textbf{Ours}    & \textbf{6.67} & 3.33 & \textbf{42.5} & \textbf{29.0} & \textbf{65.78} \\
\hline
\multirow{6}{*}{\textbf{LLaMA-3.1-8B}}
& Greedy           & 6.7  & \textbf{10.0} & 20.0 & 13.0 & 46.4 \\
& Maj@8            & 0.0  & \textbf{10.0} & 22.5 & 20.0 & 49.8 \\
& RM@8             & 6.7  & \textbf{10.0} & 25.0 & \textbf{24.0} & 51.0 \\
& Beam ($B=8$)     & \textbf{10.0} & 6.7 & 25.0 & 18.5 & 49.2 \\
& Lookahead        & 6.7  & 6.7 & 25.0 & 18.5 & 48.8 \\
& \textbf{Ours}    & 6.7   & 6.7  & \textbf{32.5} & 22.5 & \textbf{56.2} \\
\hline
\end{tabular}
\end{table*}
\subsection{Rollout Initialization}
\label{sec:method:init}
 
To promote diversity, the $N$ initial reasoning traces are sampled independently from the generator policy at a fixed sampling temperature $\Tsample > 0$, i.e., $r_i^{(0)} \stackrel{\text{i.i.d.}}{\sim} \polG(\cdot \mid x)$ for $i=1,\ldots,N$, where each draw is conditioned only on the input problem $x$, and $r_i^{(0)}$ denotes the initial depth-0 reasoning trace for rollout $i$. This breadth component directly counteracts the sampling-redundancy phenomenon identified in Section~\ref{sec:motivation}.
 
\subsection{Iterative Self-Critique and Self-Correction}\label{sec:method:refine}

For each refinement depth $d \in \{1,\ldots,D\}$, every rollout
$i \in \{1,\ldots,N\}$ is updated through a three-stage refinement
cycle that maps the previous corrected trace $r_i^{(d-1)}$ to a new
corrected trace $r_i^{(d)}$.

\paragraph{Reasoning continuation.}
The generator first produces an intermediate reasoning trace
$g_i^{(d)}$ by extending or rewriting the previous corrected trace
$r_i^{(d-1)}$ conditioned on the input problem:
\begin{equation}
  g_i^{(d)}
  \sim
  \polG\!\left(\cdot \mid x, r_i^{(d-1)}\right).
  \label{eq:reasoning}
\end{equation}

\paragraph{Self-critique.}
The critic then examines $g_i^{(d)}$ and produces a natural-language
critique $c_i^{(d)}$ identifying possible logical, arithmetic, or
structural errors:
\begin{equation}
  c_i^{(d)}
  \sim
  \polC\!\left(\cdot \mid x, g_i^{(d)}\right).
  \label{eq:critique}
\end{equation}
When no error is detected, the critique $c_i^{(d)}$ explicitly
confirms the trace, allowing the correction step below to preserve
the current solution.

\paragraph{Self-correction.}
Finally, the corrector revises the reasoning trace, producing the
depth-$d$ corrected trace $r_i^{(d)}$ conditioned on both the
intermediate reasoning $g_i^{(d)}$ and its critique $c_i^{(d)}$:
\begin{equation}
  r_i^{(d)}
  \sim
  \polR\!\left(\cdot \mid x, g_i^{(d)}, c_i^{(d)}\right).
  \label{eq:correction}
\end{equation}
The corrected trace $r_i^{(d)}$ is then used as the input context
for the next refinement depth $d{+}1$ (or, when $d = D$, as the
terminal trace passed to answer extraction).

\subsection{Answer Extraction and Aggregation} \label{sec:method:agg}
After $D$ refinement steps, each rollout $i$ yields a terminal corrected trace $r_i^{(D)}$, from which the deterministic extractor $\Extract(\cdot)$ produces the per-rollout candidate answer $\hat{a}_i = \Extract\!\bigl(r_i^{(D)}\bigr)$ for $i=1,\ldots,N$, where $\hat{a}_i \in \mathcal{Y}$. Here, $\mathcal{Y}$ denotes the normalized task-specific answer space used for evaluation. The final prediction $\hat{a}\in\mathcal{Y}$ is obtained by plurality voting over all $N$ per-rollout candidates, i.e., $\hat{a}=\Vote\!\bigl(\{\hat{a}_i\}_{i=1}^{N}\bigr)=\argmax_{a\in\mathcal{Y}}\sum_{i=1}^{N}\Ind[\hat{a}_i=a]$. Crucially, this requires no reward model, no verifier score, and no learned selection policy; $\Vote(\cdot)$ is a parameter-free aggregation operator applied to the outputs of the refinement pipeline. 
 
 \begin{figure*}[t]
  \centering
  \includegraphics[width=0.95\textwidth]{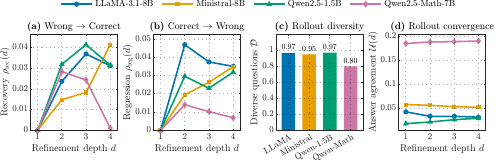}
\caption{\textbf{Refinement dynamics across depths.} Panels show recovery, regression, terminal rollout diversity, and answer agreement during refinement.}
  \label{fig:recovery_diversity}
\end{figure*}

\subsection{Compute Budget and TFLOP Accounting}\label{sec:method:budget}
We estimate inference compute using the standard transformer scaling-law approximation~\cite{kaplan2020scaling, hoffmann2022training}. For a method $m$, let $\mathcal{G}_m$ denote the set of autoregressive generation calls it performs, and let each call $u\in\mathcal{G}_m$ have realized input and output lengths ${\tin}_u$ and ${\tout}_u$. Let $\Pgen$ and $\Prm$ denote the parameter counts of the base generator model and the reward model, respectively. The generator-side compute is $\Compute^{\mathrm{gen}}_m=\sum_{u\in\mathcal{G}_m}2\Pgen\bigl({\tin}_u+{\tout}_u\bigr)$. Verifier-based methods additionally perform prefill-only reward-model scoring: if $\mathcal{R}_m$ is the set of reward-model scoring calls and ${\tin}^{\mathrm{rm}}_v$ is the realized input length for scoring call $v$, then $\Compute^{\mathrm{rm}}_m=\sum_{v\in\mathcal{R}_m}2\Prm{\tin}^{\mathrm{rm}}_v$. The total inference compute is therefore $\Compute_m=\Compute^{\mathrm{gen}}_m+\Compute^{\mathrm{rm}}_m$, with $\Compute^{\mathrm{rm}}_m=0$ for methods that do not use an external verifier.

Best-of-$N$ sampling generates $N$ independent candidate solutions. Our method instead refines each of the $N$ rollouts through repeated reasoning continuation, self-critique, and self-correction steps, all using the same base model $\polTheta$ and no reward model. RM@$N$ uses the same sampled candidates as Best-of-$N$, but adds reward-model scoring for final selection.

All token counts are measured from the realized prompts and responses after the same truncation rules used during inference, so the reported totals reflect the actual contexts processed by each model. Figure~\ref{fig:compute_tflops_app} reports the resulting TFLOP totals across methods, models, and benchmarks; Table~\ref{tab:compute_normalized_by_model} uses these totals to compare accuracy gains against the additional compute required by our refinement procedure; and Section~\ref{sec:results:dynamics_compute} discusses the resulting refinement dynamics and compute-normalized trade-offs.

\begin{table}[t]
\centering
% \small
\caption{\textbf{Ablation of the self-critique stage} ($N{=}8$, $D{=}4$). \textbf{w/o}: corrector receives only the reasoning trace; \textbf{w/}: full pipeline with explicit critique.}
\label{tab:self_critic_ablation}
\resizebox{\columnwidth}{!}{%
\begin{tabular}{lcccc}
\toprule
\multirow{2}{*}{\textbf{Dataset}}
& \multicolumn{2}{c}{\textbf{Qwen2.5-Math-7B}}
& \multicolumn{2}{c}{\textbf{Qwen2.5-1.5B}} \\
\cmidrule(lr){2-3}\cmidrule(lr){4-5}
& \textbf{w/o} & \textbf{w/}
& \textbf{w/o} & \textbf{w/} \\
\midrule
AIME24        & 10.0 & 10.0  & 3.3  & 6.67 \\
AIME25        & 13.3 & 16.67 & 0.0  & 6.67 \\
AMC           & 65.0 & 67.5  & 27.5 & 32.5 \\
OlyBench      & 44.9 & 47.2  & 23.0 & 24.5 \\
MATH          & 80.4 & 81.6  & 55.6 & 58.0 \\
\bottomrule
\end{tabular}%
}
\end{table}
% We evaluate on five mathematical reasoning benchmarks spanning a broad difficulty range: \textbf{AIME24} and \textbf{AIME25} (competition-level olympiad problems, 30 problems each), \textbf{AMC} (40 problems), \textbf{OlympiadBench (OlyBench)} (674 competition problems)~\cite{he2024olympiadbench}, and \textbf{MATH500} (500 held-out problems)~\cite{hendrycks2021measuring}. Accuracy is reported as the percentage of questions whose majority-vote answer matches the gold answer after string normalization. We test four open-weight models that span a wide capability range: Qwen2.5-Math-7B-Instruct~\cite{yang2024qwen25mathtechnicalreportmathematical}, Qwen2.5-1.5B~\cite{qwen2}, Ministral-8B~\cite{ministral}, and LLaMA-3.1-8B~\cite{grattafiori2024llama}. Unless stated otherwise, all runs use $N{=}8$ rollouts, depth $D{=}4$, and rollout sampling temperature $\Tsample{=}0.7$.

\section{Experiments} \label{sec:experiments}
\subsection{Benchmarks and Models} \label{sec:exp:setup}
We evaluate on five mathematical reasoning benchmarks spanning a broad difficulty range: \textbf{AIME24}, \textbf{AIME25}, \textbf{AMC}, \textbf{OlympiadBench (OlyBench)}~\cite{he2024olympiadbench}, and \textbf{MATH500}~\cite{hendrycks2021measuring}. Accuracy is reported as the percentage of questions whose majority-vote answer matches the gold answer after string normalization.

We test four open-weight models that span a wide capability range: Qwen2.5-Math-7B-Instruct~\cite{yang2024qwen25mathtechnicalreportmathematical}, Qwen2.5-1.5B~\cite{qwen2}, Ministral-8B~\cite{ministral}, and LLaMA-3.1-8B~\cite{grattafiori2024llama}. Unless stated otherwise, all runs
use $N{=}8$, $D{=}4$, sampling temperature $\Tsample{=}0.7$, and a single
NVIDIA H100 GPU with 80GB of memory. In addition to accuracy, we report TFLOPs and compute-normalized gains to measure inference cost and efficiency; the compute accounting is described in Section~\ref{sec:method:budget}, with aggregate results reported in Figure~\ref{fig:compute_tflops_app} and Table~\ref{tab:compute_normalized_by_model}. Section~\ref{sec:results:dynamics_compute} analyzes recovery, regression, rollout diversity, and answer agreement across refinement depths, with metric definitions in Appendices~\ref{app:metrics:recov} and~\ref{app:metrics:div}.

\subsection{Baselines} \label{sec:exp:baselines}
We compare against five inference-time baselines, all using the same base model without additional training. \textbf{Greedy} decoding produces one deterministic solution with $\Tsample=0$. \textbf{Majority Vote (Maj@8)} samples eight independent solutions and selects the most frequent answer~\cite{wang2022self}. \textbf{Best-of-$N$ with Verifier (RM@8)} samples the same eight solutions but selects the one with the highest Qwen2.5-Math-RM-32B score~\cite{yang2024qwen25math}. \textbf{Beam Search} uses token-level beam search with width $B=8$; we include it as a standard decoding baseline, while noting that it is weaker for chain-of-thought reasoning because it does not search at the solution level. \textbf{Lookahead} samples $N=8$ continuations at each of $K=3$ checkpoints and selects the highest-confidence path. All sampling-based baselines use $N=8$ candidates to keep the primary sample budget fixed; our main equal-sample comparisons are therefore Maj@8 and RM@8.
\begin{figure*}[t]
  \centering
  \includegraphics[width=\textwidth]{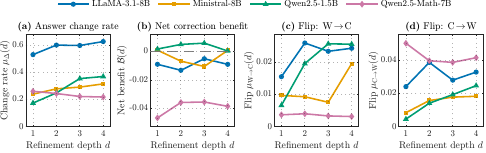}
  \caption{\textbf{Correction behavior across depths.} Panels report answer changes, net correction benefit, and wrong-to-correct versus correct-to-wrong flips during refinement.}
  \label{fig:correction_effectiveness}
\end{figure*}

\begin{table}[tb]
\centering
% \small
\setlength{\tabcolsep}{4pt}
\caption{AMC23 accuracy (\%) under varying breadth--depth allocations.}
\label{tab:amc23_nd_sweep}
\begin{tabular}{@{}lccccc@{}}
\toprule
\textbf{Model}
  & $N$
  & $D\!=\!2$
  & $D\!=\!3$
  & $D\!=\!4$
  & $D\!=\!5$ \\
\midrule
\multirow{4}{*}{Qwen2.5-1.5B}
  & 3  & 21.05 & 27.5 & 20.5 & 20.5 \\
  & 5  & 17.5  & 25.0 & 32.5 & 32.5 \\
  & 8  & 25.0  & 27.5 & 32.5 & 27.5 \\
  & 10 & 32.5  & 32.5 & 32.5 & 27.5 \\
\addlinespace[3pt]
\multirow{4}{*}{Ministral-8B}
  & 3  & 25.0 & 32.5 & 22.5 & 37.5 \\
  & 5  & 37.5 & 42.5 & 42.5 & 37.5 \\
  & 8  & 45.0 & 35.0 & 42.5 & 40.0 \\
  & 10 & 42.5 & 40.0 & 42.5 & 37.5 \\
\bottomrule
\end{tabular}
\end{table}

\section{Results}\label{sec:results}
We evaluate whether breadth--depth refinement improves reasoning accuracy, whether the self-critic stage is necessary, and how the gains trade off against additional inference compute. Table~\ref{tab:selected_math_results_transposed} reports the main results. Overall, our method improves across model scales, with the largest gains on benchmarks where refinement can repair multi-step reasoning errors before aggregation.

For Qwen2.5-Math-7B, refinement improves over the strongest baseline on AIME25, AMC, OlympiadBench, and MATH500. The gains are larger for Qwen2.5-1.5B, especially on MATH500, where accuracy increases from 29.6\% with RM@8 to 58.0\%; Ministral-8B also improves clearly on AMC and MATH500. These results suggest that refinement is most useful when the base model has latent reasoning ability that is not reliably expressed in a single sampled solution. All gains are obtained using only the base model and majority voting over refined rollouts, without an external verifier, PRM, learned stopping policy, or additional training. We note that the main comparison fixes the primary sample count rather than the total TFLOPs. The compute-normalized gain $\eta$, defined in Appendix~\ref{app:compute_tflops}, measures accuracy improvement per additional TFLOP, but does not constitute a direct equal-budget comparison in which majority voting is scaled to the same compute.

\subsection{Self-Critic Ablation}
\label{sec:results:ablation}
Table~\ref{tab:self_critic_ablation} shows that the explicit self-critic stage improves refinement in most settings, indicating that critique provides useful intermediate feedback rather than merely increasing the token budget. The effect is most pronounced for Qwen2.5-1.5B: adding the critic improves AIME25 from $0.0\%$ to $6.67\%$, AMC from $27.5\%$ to $32.5\%$, and MATH500 from $55.6\%$ to $58.0\%$. Qwen2.5-Math-7B shows smaller but consistent gains on several benchmarks, such as AMC from $65.0\%$ to $67.5\%$ and OlympiadBench from $44.9\%$ to $47.2\%$. These results suggest that explicit critique is especially useful for weaker models, where first-pass reasoning is less stable and correction benefits from a more structured error signal.

\subsection{Refinement Dynamics and Compute Trade-off}
\label{sec:results:dynamics_compute}
We next analyze how answers evolve across refinement depth on the same evaluation suite used in Table~\ref{tab:selected_math_results_transposed}. Figure~\ref{fig:recovery_diversity} reports four diagnostic metrics: recovery measures wrong-to-correct majority changes, regression measures correct-to-wrong majority changes, rollout diversity measures whether refined trajectories remain distinct, and answer agreement measures convergence among rollouts. Refinement continues to produce recoveries across depths while keeping regressions limited. Diversity also remains high through depth~$4$ ($0.97$ for LLaMA-3.1-8B, $0.95$ for Ministral-8B, $0.97$ for Qwen2.5-1.5B, and $0.80$ for Qwen2.5-Math-7B), showing that refinement preserves breadth rather than collapsing all rollouts to the same answer. 

Figure~\ref{fig:correction_effectiveness} analyzes single-rollout correction behavior: answer-change rate measures how often refinement changes a prediction, net correction benefit measures whether these changes are helpful overall, and wrong-to-correct versus correct-to-wrong flips separate useful repairs from harmful over-corrections. These results show that individual corrections are noisy, so majority voting helps suppress isolated regressions while retaining improvements shared across refined rollouts. Formal definitions of recovery and regression are given in Appendix~\ref{app:metrics:recov}, diversity and agreement are defined in Appendix~\ref{app:metrics:div}, and Appendix~\ref{app:example-sec-box} provides a qualitative refinement trace.

\subsection{Compute and Breadth-Depth Trade-off}
\label{sec:results:compute}
Refinement incurs additional inference compute because each rollout is processed through reasoning, continuation, critique, and correction; we therefore report total TFLOPs in Figure~\ref{fig:compute_tflops_app} and compute-normalized gains in Table~\ref{tab:compute_normalized_by_model} and Figure~\ref{fig:compute_normalized_gain}. Compute-normalized gain measures the accuracy improvement obtained per unit of additional compute, with the formal definition given in Appendix~\ref{app:compute_tflops}. The gains are positive across models and baselines, but are largest for Qwen2.5-1.5B: compared with Maj@8, refinement improves average accuracy from $13.21\%$ to $25.67\%$ with $\eta=17.10$, and compared with RM@8 it improves from $14.95\%$ to $25.67\%$ with $\eta=17.01$. For stronger models, the gains remain positive but smaller; for example, Qwen2.5-Math-7B improves over RM@8 by $2.36$ average accuracy points with $\eta=1.16$, suggesting that verifier and search baselines already recover part of the available headroom. Table~\ref{tab:amc23_nd_sweep} further studies the breadth--depth allocation on AMC23 and shows that accuracy is not monotonic in either $N$ or $D$: Qwen2.5-1.5B reaches $32.5\%$ under several moderate settings, while Ministral-8B ranges from $42.5\%$ at $(N,D)=(5,3)$ or $(5,4)$ to $45.0\%$ at $(8,2)$. Overall, these results support using moderate breadth and depth, with $N=8$ and $D=4$ providing a stable trade-off between preserving diverse reasoning paths and allowing enough refinement depth to repair errors before aggregation.

\section{Related Work}

\paragraph{Test-time scaling and reasoning aggregation.}
Test-time scaling improves LLM reasoning by allocating additional inference compute through repeated sampling, search, branching, reranking, or optimization without updating model parameters~\cite{beirami2024theoretical, zuo2026ttrl, inoue2026wider, wang2026nablareasoner, li2025reasoning}. Self-consistency is a standard instance of this paradigm, aggregating multiple sampled reasoning traces through majority voting~\cite{wang2022self}. Ranked voting based self-consistency further improves aggregation by using ranked candidate answers rather than only the most frequent top answer~\cite{wang2025ranked}. Other approaches expand the reasoning space through tree search or trajectory optimization~\cite{song2024trial, xie2024human, zhang2024rest}. However, these methods often aggregate or search over sampled trajectories without explicitly improving each trajectory across refinement depths.

\paragraph{Verifier-guided reasoning.}
Verifier-based methods select or guide candidate solutions using outcome reward models or process reward models~\cite{cobbe2021training, lightman2023let, li2023making, wang2024math}. MindStar improves mathematical reasoning at inference time by using a process reward model to guide search over reasoning steps~\cite{kang2024mindstar}. Although such verifiers can improve mathematical reasoning, their effectiveness depends on the quality and calibration of the scoring model, and recent work highlights limitations of reward-model-based selection and inference-time verification~\cite{dorner2025roc, li2025fixing, zhang2025lessons}. This dependence can make verifier-guided reasoning less reliable when the verifier is miscalibrated or unavailable.

\paragraph{Self-critique and self-refinement.}
Self-refinement methods improve outputs by asking the model to critique and revise its own generations. Self-Refine introduces an iterative feedback-and-revision loop~\cite{madaan2023selfrefine}, Reflexion uses verbal feedback to improve later attempts~\cite{shinn2024reflexion}, and Monte Carlo Tree Self-Refine searches over refinement trajectories~\cite{zhang2024accessing}. Multi-agent reflection and progressive refinement further show that iterative feedback can improve reasoning trajectories~\cite{yuan2025reinforce, du2025think}. However, individual refinement trajectories can be noisy and may introduce new errors during correction.

\paragraph{Intrinsic self-correction and self-improvement.}
Intrinsic self-correction remains difficult: LLMs can fail to detect their own reasoning errors and may even degrade after naive correction~\cite{huang2024large}. Prior work therefore studies explicit mechanisms for self-correction and self-improvement, including self-correcting sequence generation~\cite{welleck2022generating}, STaR-style bootstrapping from generated rationales~\cite{zelikman2022star}, implicit or search-based self-improvement~\cite{tian2024toward, wang2024enabling}, and analyses of self-improvement reversal in iterative post-training~\cite{wu2025progress}. More recent methods train models for correction or verification: SCoRe uses multi-turn reinforcement learning for intrinsic self-correction~\cite{kumar2024training}, while ReVISE learns an intrinsic stop-or-refine mechanism for test-time correction~\cite{lee2025revise}. Such approaches may require additional training, supervision, or learned correction mechanisms.

\section{Conclusion}
Width-only test-time scaling can suffer from diminishing returns because additional samples often repeat dominant answer patterns rather than provide new reasoning evidence. Verifier-based selection can help, but it introduces dependence on reward-model calibration. We instead use test-time compute to refine sampled trajectories before aggregation. Our breadth--depth framework preserves multiple reasoning directions while iteratively repairing local errors through self-critique and self-correction, then aggregates the refined outputs by majority voting without any external verifier, learned stopping policy, or additional training. Useful reasoning evidence comes not only from sampling many attempts but also from allowing each attempt to repair unstable intermediate steps before aggregation. This makes the method robust to noisy individual corrections, since final decisions are based on agreement across refined rollouts rather than a single trajectory or verifier score. More broadly, this suggests a path toward future test-time scaling methods that use additional compute not only to search wider but also to refine and stabilize reasoning before aggregation.

\section*{Limitations}
This work has three main limitations. First, the proposed method has higher inference cost because each rollout goes through reasoning, critique, and correction steps. Although batching improves throughput, the method still requires more forward passes than greedy decoding, majority voting, or standard best-of-$N$ sampling. Appendix~\ref{app:compute_tflops} and Appendix~\ref{app:refinement_compute_diagnostics} report compute-normalized results that account for this additional TFLOP cost. Second, the approach depends on the base model's ability to critique and revise its own reasoning. If the model produces an inaccurate critique or over-corrects a valid solution, refinement can introduce rollout-level regressions, even though majority voting helps reduce their effect. Third, our evaluation focuses on mathematical reasoning benchmarks, where answers are well-defined and extraction is relatively reliable. The effectiveness of verifier-free self-refinement may vary in domains with more open-ended outputs, ambiguous evaluation criteria, or weaker self-critique signals.
\bibliography{custom}

\appendix

\section{Diagnostic Metric Definitions}
\label{app:metrics}

This appendix defines the diagnostic metrics used to analyze refinement dynamics, rollout diversity, sampling redundancy, and compute-normalized efficiency. These metrics support the analysis in Section~\ref{sec:results:dynamics_compute}, the redundancy motivation in Figure~\ref{fig:motivation}, the refinement dynamics in Figure~\ref{fig:recovery_diversity}, the correction behavior in Figure~\ref{fig:correction_effectiveness}, and the compute comparisons in Figure~\ref{fig:compute_tflops_app}, Table~\ref{tab:compute_normalized_by_model}, and Figure~\ref{fig:compute_normalized_gain}.

\paragraph{Shared notation.}
Let $\Qset$ denote the set of evaluation questions and let $d \in \{1,\ldots,D\}$ denote the refinement depth. For question $q$ with gold answer $a_q^{\star}$, the depth-$d$ plurality-vote answer is
\[
  \hat{a}_q^{(d)}
  =
  \Vote\!\bigl(\{\Extract(r_{q,i}^{(d)})\}_{i=1}^{N}\bigr),
\]
where this quantity is used only for analysis at intermediate depths $d<D$, while the method itself uses only the terminal vote $\hat{a}_q^{(D)}$. We define the correctness indicator as $\delta_q(d) \triangleq \Ind[\Norm(\hat{a}_q^{(d)})=\Norm(a_q^{\star})]$, where $\Norm(\cdot)$ is the answer-normalization map defined in Appendix~\ref{app:metrics:norm}. For rollout-level correction analysis, let $\alpha_{q,i}^{(d)} \triangleq \Extract(g_{q,i}^{(d)})$ and $\beta_{q,i}^{(d)} \triangleq \Extract(r_{q,i}^{(d)})$ denote the pre-correction and post-correction answers for rollout $i$ at depth $d$.

\subsection{Depth-Wise Accuracy and Improvement Rate}
\label{app:metrics:acc}

Depth-wise accuracy measures the plurality-vote accuracy after each refinement depth:
\[
  \mathcal{A}(d)
  \triangleq
  \frac{1}{|\Qset|}
  \sum_{q \in \Qset}
  \delta_q(d),
  \qquad
  d \in \{1,\ldots,D\}.
\]
The depth-wise improvement rate measures the incremental change in accuracy between two consecutive refinement depths:
\[
  \Delta\mathcal{A}(d)
  \triangleq
  \mathcal{A}(d)-\mathcal{A}(d-1).
\]
For $d=1$, $\mathcal{A}(0)$ denotes the plurality-vote accuracy before refinement, computed from the initial rollouts $\{r_{q,i}^{(0)}\}_{i=1}^{N}$.

\subsection{Recovery and Regression Rates}
\label{app:metrics:recov}

Recovery measures questions whose plurality-vote answer changes from incorrect to correct after one refinement step, while regression measures questions whose plurality-vote answer changes from correct to incorrect. For $d \geq 1$, these rates are defined as
\begin{align}
  \rho_{\mathrm{rec}}(d)
  &=
  \Expect_{q \sim \Qset}
  \bigl[
    \Ind[\delta_q(d-1)=0 \;\wedge\; \delta_q(d)=1]
  \bigr],
  \label{eq:recovery-rate}\\
  \rho_{\mathrm{reg}}(d)
  &=
  \Expect_{q \sim \Qset}
  \bigl[
    \Ind[\delta_q(d-1)=1 \;\wedge\; \delta_q(d)=0]
  \bigr].
  \label{eq:regression-rate}
\end{align}
These quantities explain the net change in depth-wise accuracy:
\[
  \Delta\mathcal{A}(d)
  =
  \rho_{\mathrm{rec}}(d)
  -
  \rho_{\mathrm{reg}}(d).
\]
Thus, refinement improves aggregate accuracy when recoveries outweigh regressions. These rates are visualized in Figure~\ref{fig:recovery_diversity}.

\subsection{Correction Effectiveness}
\label{app:metrics:corr}

Correction effectiveness measures how often the corrector changes individual rollout answers and whether those changes move the answer toward or away from the gold answer. Let
\[
  \Sset_d
  =
  \{(q,i): q\in\Qset,\; i\in[N]\}
\]
be the set of all question--rollout pairs at depth $d$. We define the answer-change indicator as
\[
  \chi_{q,i}^{(d)}
  \triangleq
  \Ind\!\left[
    \Norm(\alpha_{q,i}^{(d)})
    \neq
    \Norm(\beta_{q,i}^{(d)})
  \right].
\]
The answer-change rate is
\[
  \mu_{\Delta}(d)
  \triangleq
  \Expect_{(q,i)\sim\Sset_d}
  \left[
    \chi_{q,i}^{(d)}
  \right].
\]
To measure the direction of these changes, we define the pre-correction and post-correction correctness indicators as
\[
\begin{aligned}
  \rho_{q,i}^{(d)}
  &\triangleq
  \Ind\!\left[
    \Norm(\alpha_{q,i}^{(d)})
    =
    \Norm(a_q^{\star})
  \right], \\
  \kappa_{q,i}^{(d)}
  &\triangleq
  \Ind\!\left[
    \Norm(\beta_{q,i}^{(d)})
    =
    \Norm(a_q^{\star})
  \right].
\end{aligned}
\]
The wrong-to-correct and correct-to-wrong flip rates are
\begin{align}
  \mu_{\mathsf{W}\to\mathsf{C}}(d)
  &\triangleq
  \Expect_{\Sset_d}
  \Bigl[
    \chi_{q,i}^{(d)}
    \Ind\!\bigl[
      \rho_{q,i}^{(d)}=0
      \wedge
      \kappa_{q,i}^{(d)}=1
    \bigr]
  \Bigr],
  \label{eq:wrong-to-correct} \\
  \mu_{\mathsf{C}\to\mathsf{W}}(d)
  &\triangleq
  \Expect_{\Sset_d}
  \Bigl[
    \chi_{q,i}^{(d)}
    \Ind\!\bigl[
      \rho_{q,i}^{(d)}=1
      \wedge
      \kappa_{q,i}^{(d)}=0
    \bigr]
  \Bigr].
  \label{eq:correct-to-wrong}
\end{align}
The net correction benefit is
\[
  \mathcal{B}(d)
  \triangleq
  \mu_{\mathsf{W}\to\mathsf{C}}(d)
  -
  \mu_{\mathsf{C}\to\mathsf{W}}(d).
\]
A negative $\mathcal{B}(d)$ at the rollout level does not necessarily imply lower final accuracy, because the method aggregates $N$ refined rollouts by plurality voting. Thus, isolated harmful flips can be suppressed when they are not consistent across rollouts, while repeated beneficial flips can shift the final vote. These quantities are reported in Figure~\ref{fig:correction_effectiveness}.

\subsection{Rollout Diversity and Agreement}
\label{app:metrics:div}

Rollout diversity measures whether refinement preserves multiple candidate answers across the $N$ rollouts instead of collapsing all trajectories to the same answer. For question $q$ at depth $d$, define the set of distinct non-empty normalized answers as
\[
  \Aset_q^{(d)}
  \triangleq
  \Bigl\{
    \Norm(\Extract(r_{q,i}^{(d)}))
    :
    i\in[N],
    \;
    \Norm(\Extract(r_{q,i}^{(d)}))\neq\varnothing
  \Bigr\}.
\]
The terminal diversity rate is
\[
  \mathcal{D}
  \triangleq
  \Expect_{q\sim\Qset}
  \bigl[
    \Ind[|\Aset_q^{(D)}|>1]
  \bigr],
\]
which measures the fraction of questions for which at least two distinct terminal answers remain after refinement. The answer-agreement rate is
\[
  \mathcal{U}(d)
  \triangleq
  \Expect_{q\sim\Qset}
  \bigl[
    \Ind[|\Aset_q^{(d)}|=1]
  \bigr],
\]
which measures the fraction of questions for which all non-empty rollout answers agree at depth $d$. High $\mathcal{D}$ and low $\mathcal{U}(d)$ indicate that refinement preserves breadth, leaving meaningful disagreements for plurality voting to resolve. These metrics are shown in Figure~\ref{fig:recovery_diversity}.

\subsection{Answer Normalization}
\label{app:metrics:norm}

The normalization map $\Norm(\cdot)$ is applied before comparing predicted and gold answers. It strips surrounding whitespace, removes common \LaTeX\ spacing tokens such as \texttt{\textbackslash,}, \texttt{\textbackslash!}, and \texttt{\textbackslash;}, unwraps one layer of enclosing curly braces, removes trailing decimal zeros where applicable, and lowercases the resulting string. For example, $3.500$ is normalized to $3.5$. This lightweight normalization is used consistently for answer extraction, voting diagnostics, recovery/regression analysis, and rollout-level correction metrics.

\subsection{Sampling Redundancy Metrics}
\label{app:metrics:redundancy}

The redundancy metrics quantify whether additional initial samples produce genuinely new reasoning directions or mainly repeat existing answer patterns. They are used for the motivation analysis in Figure~\ref{fig:motivation}. For an input problem $x$, let $\{r_i^{(0)}\}_{i=1}^{N}$ be the $N$ initial rollouts sampled from the generator policy. Each rollout is mapped to a unit-normalized sentence embedding,
\[
  e_i
  =
  \frac{\phi(r_i^{(0)})}{\|\phi(r_i^{(0)})\|_2}
  \in
  \mathbb{S}^{\embdim-1},
\]
where $\phi$ is the \texttt{all-MiniLM-L6-v2} sentence encoder with $\embdim=384$ \cite{reimers-2019-sentence-bert}. We cluster the embeddings using agglomerative clustering with average linkage and cosine distance, using threshold $1-\simthr$ with $\simthr=0.85$. This yields reasoning-direction clusters $\{\mathcal{C}_j\}_{j\geq1}$, with empirical cluster mass $p_j(N)\triangleq \frac{1}{N}\sum_{i=1}^{N}\Ind[r_i^{(0)}\in\mathcal{C}_j]$. We report $\mathcal{R}(N)\triangleq |\{j:\mathcal{C}_j\neq\varnothing\}|$, $\mathcal{R}(N)/N$, $f_{\max}(N)\triangleq \max_j p_j(N)$, and $\mathcal{H}(N)\triangleq -\sum_j p_j(N)\log_2 p_j(N)$, which measure the number of unique reasoning directions, diversity yield per sample, dominant-cluster concentration, and cluster entropy, respectively. A decreasing $\mathcal{R}(N)/N$ and slowly growing $\mathcal{H}(N)$ indicate redundancy saturation: larger sampling budgets increasingly revisit existing directions rather than expanding the reasoning space. All quantities are averaged per problem over the benchmark, using Qwen2.5-Math-7B-Instruct with $\Tsample=0.7$ and fixed $\simthr=0.85$.

\paragraph{Robustness of the clustering analysis.}
The reported statistics use $\simthr=0.85$ and the \texttt{all-MiniLM-L6-v2} sentence encoder applied to full reasoning traces. The analysis uses answer-based clustering as the primary grouping (each rollout is assigned by its extracted normalized final answer), which is deterministic given the answer extractor and does not depend on the sentence embedding at all. The sentence-embedding clustering is used only for the redundancy analysis in Figure~\ref{fig:motivation} and does not affect the main accuracy results.
\begin{figure*}[t]
  \centering
  \includegraphics[width=\textwidth]{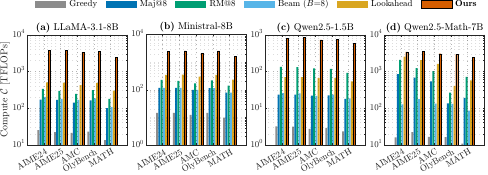}
  \caption{Total inference compute in TFLOPs, shown on a log scale, for each method and benchmark across the four model backbones.}
  \label{fig:compute_tflops_app}
\end{figure*}
\subsection{Compute Accounting and Compute-Normalized Gain}
\label{app:compute_tflops}

Compute is estimated using the transformer FLOP accounting described in Section~\ref{sec:method:budget}. To compare methods on a common efficiency axis, we define compute-normalized gain as
\begin{equation}
  \CNG
  \triangleq
  \frac{
    10^{3}\bigl(a^{\mathrm{ours}}-a^{\mathrm{base}}\bigr)
  }{
    \Compute^{\mathrm{ours}}-\Compute^{\mathrm{base}}
  } .
  \label{eq:cng}
\end{equation}
Here, $a$ denotes the average accuracy across the five benchmarks, and $\Compute$ denotes the corresponding average inference compute. Larger $\CNG$ indicates that the method obtains more accuracy improvement per additional $10^{3}$ TFLOPs. Figure~\ref{fig:compute_tflops_app} reports the total TFLOPs per method and benchmark, while Table~\ref{tab:compute_normalized_by_model} reports the resulting compute-normalized comparison against each baseline.

\begin{figure*}[t]
  \centering
  \includegraphics[width=\textwidth]{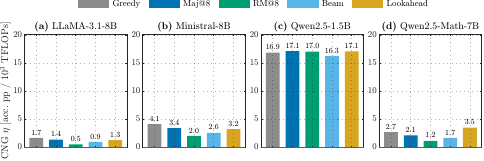}
  \caption{\textbf{Compute-normalized gain of structured refinement over each baseline.} Bars report average accuracy-point gain per additional $10^{3}$ TFLOPs across the five benchmarks.}
  \label{fig:compute_normalized_gain}
\end{figure*}
\begin{table*}[t]
\centering
\small
\caption{\textbf{Compute-normalized comparison across baselines.} $\Delta$Acc.\ denotes the average accuracy gain of our method over each baseline, $\Delta$TFLOPs denotes the additional compute used, and $\CNG$ reports accuracy-point gain per additional $10^{3}$ TFLOPs. W/T/L counts benchmark-level wins, ties, and losses.}
\label{tab:compute_normalized_by_model}
\begin{tabular}{llcccccc}
\hline
\textbf{Model} & \textbf{Baseline}
& \textbf{Base Acc.}
& \textbf{Ours Acc.}
& $\boldsymbol{\Delta}$\textbf{Acc.}
& $\boldsymbol{\Delta}$\textbf{TFLOPs}
& $\boldsymbol{\CNG}$
& \textbf{W/T/L} \\
\hline
\multirow{5}{*}{\textbf{Qwen2.5-Math-7B}}
& Greedy       & 36.33 & 44.59 & 8.27 & 3089.09 & 2.68 & 5/0/0 \\
& Maj@8        & 39.03 & 44.59 & 5.56 & 2624.06 & 2.12 & 4/1/0 \\
& RM@8         & 42.23 & 44.59 & 2.36 & 2033.84 & 1.16 & 4/0/1 \\
& Beam ($B=8$) & 39.60 & 44.59 & 4.99 & 2973.83 & 1.68 & 4/1/0 \\
& Lookahead    & 38.79 & 44.59 & 5.80 & 1661.06 & 3.49 & 4/1/0 \\
\hline
\multirow{5}{*}{\textbf{Qwen2.5-1.5B}}
& Greedy       & 13.07 & 25.67 & 12.59 & 747.54 & 16.85 & 4/1/0 \\
& Maj@8        & 13.21 & 25.67 & 12.46 & 728.39 & 17.10 & 5/0/0 \\
& RM@8         & 14.95 & 25.67 & 10.72 & 629.85 & 17.01 & 5/0/0 \\
& Beam ($B=8$) & 13.83 & 25.67 & 11.84 & 727.39 & 16.27 & 5/0/0 \\
& Lookahead    & 13.99 & 25.67 & 11.68 & 684.31 & 17.06 & 5/0/0 \\
\hline
\multirow{5}{*}{\textbf{Ministral-8B}}
& Greedy       & 20.03 & 29.46 & 9.43 & 2277.12 & 4.14 & 5/0/0 \\
& Maj@8        & 22.04 & 29.46 & 7.42 & 2183.71 & 3.40 & 5/0/0 \\
& RM@8         & 25.18 & 29.46 & 4.28 & 2094.22 & 2.04 & 3/2/0 \\
& Beam ($B=8$) & 23.81 & 29.46 & 5.64 & 2183.71 & 2.58 & 4/0/1 \\
& Lookahead    & 23.11 & 29.46 & 6.34 & 1970.19 & 3.22 & 4/1/0 \\
\hline
\multirow{5}{*}{\textbf{LLaMA-3.1-8B}}
& Greedy       & 19.22 & 24.92 & 5.70 & 3384.67 & 1.68 & 3/1/1 \\
& Maj@8        & 20.46 & 24.92 & 4.46 & 3256.72 & 1.37 & 4/0/1 \\
& RM@8         & 23.34 & 24.92 & 1.58 & 3131.56 & 0.50 & 2/1/2 \\
& Beam ($B=8$) & 21.88 & 24.92 & 3.04 & 3236.74 & 0.94 & 3/1/1 \\
& Lookahead    & 21.14 & 24.92 & 3.78 & 2958.54 & 1.28 & 3/2/0 \\
\hline
\end{tabular}
\end{table*}

\subsection{Interpretation of Refinement and Compute Diagnostics}
\label{app:refinement_compute_diagnostics}

The recovery and regression curves in Figure~\ref{fig:recovery_diversity} show that refinement continues to repair some incorrect majority answers across depths, while harmful regressions remain limited. This supports the main intuition of the method: depth is useful because it gives each sampled trajectory repeated opportunities to correct local reasoning errors before aggregation. At the same time, terminal rollout diversity remains high, with values between $0.80$ and $0.97$ at depth $4$, and full answer agreement remains relatively rare. Thus, refinement does not simply force all rollouts into the same answer; it preserves enough breadth for majority voting to remain meaningful.

Figure~\ref{fig:correction_effectiveness} shows that correction at the individual-rollout level is noisy. The answer-change rate is highest in the early depths and generally decreases as trajectories stabilize, while the net correction benefit can be negative for some models and depths. This does not contradict the aggregate accuracy gains in Table~\ref{tab:selected_math_results_transposed}, because the final prediction is not based on a single corrected rollout. Instead, the method aggregates $N=8$ refined rollouts, allowing majority voting to reduce the effect of isolated correct-to-wrong flips and retain improvements that appear consistently across trajectories.

The compute results in Table~\ref{tab:compute_normalized_by_model} and Figure~\ref{fig:compute_normalized_gain} show that structured refinement is most compute-efficient for smaller models, where there is more room to recover latent reasoning ability at relatively low cost. For Qwen2.5-1.5B, $\CNG$ is approximately $17$ accuracy points per $10^3$ additional TFLOPs against several baselines. For stronger models, the gains remain positive but smaller; for example, Qwen2.5-Math-7B improves over RM@8 with $\CNG=1.16$, suggesting that verifier-based selection already captures part of the available headroom. Overall, the diagnostics show that the benefit of the proposed method comes from reallocating test-time compute from only sampling more candidates to refining existing trajectories before voting.

\FloatBarrier

\clearpage

\section{Qualitative Example --- Self-Critique and
Self-Correction Trace} \label{app:example-sec-box}
 
{\small We present Rollout~2 from question~18 of the MATH evaluation
set. This rollout illustrates our technique's key success mode: an incorrect
initial reasoning path is detected by the self-critique module at depth~2, and
self-correction rebuilds the solution on a valid geometric basis. The correct
answer ($x = 28^\circ$) is then maintained stably through depths~3 and~4.}
 
\begin{problembox}{Problem (MATH --- Geometry) \quad
  Gold answer:~28\textdegree \quad
  Model: Qwen2.5-Math-7B-Instruct \quad $N{=}8$,\ depth~$=4$}
\small
$\overline{BC}$ is parallel to the segment through $A$, and $AB = BC$.
The angle at vertex $A$ is labeled $124^\circ$.
What is the number of degrees represented by $x^\circ$?
\end{problembox}

\smallskip\noindent\textbf{Rollout 2 --- Full Depth Trace}

\noindent\textbf{Depth 1}
 
\begin{reasonbox}{Reasoning \quad Answer:~\cmarkw~$28^\circ$}
\small
Step 1: Identify that $\angle ABC = 124^\circ$ since $AB = BC$ implies
$\triangle ABC$ is isosceles with base angles equal.\\
Step 2: Since $\overline{BC} \| \text{segment through } A$, this means
$\angle BAC = x^\circ$ because they are alternate interior angles formed
by a transversal cutting two parallel lines.\\
Step 3: The sum of the measures of the angles in any triangle is
$180^\circ$. Therefore, we can write an equation for the sum of angles in
$\triangle ABC$:
\[ \angle ABC + \angle BAC + \angle ACB = 180^\circ \]
Step 4: Substitute the known values into the equation:
\[ 124^\circ + x^\circ + x^\circ = 180^\circ \]
Step 5: Simplify and solve for $x$:
\[ 124^\circ + 2x^\circ = 180^\circ, \quad
   2x^\circ = 56^\circ, \quad x^\circ = 28^\circ \]
Therefore, the value of $x$ is $28^\circ$. \textbf{Final answer:}
$\boxed{28}$
\end{reasonbox}
 
\begin{critiquebox}{Self-Critique}
\small
The solution appears correct. There were no identified errors in the
provided solution trace.
\end{critiquebox}
 
\begin{correctionbox}{Self-Correction \quad Corrected answer:~\cmarkw~$28^\circ$}
\small
Step 1: Identify that $\angle ABC = 124^\circ$ since $AB = BC$ implies
$\triangle ABC$ is isosceles with base angles equal.\\
Step 2: Since $\overline{BC} \| \text{segment through } A$, this means
$\angle BAC = x^\circ$ because they are \emph{corresponding} angles
formed by a transversal cutting two parallel lines.\\
Step 3--5: (same as reasoning) $\Rightarrow x = 28^\circ$.
\textbf{Final answer:} $\boxed{28}$
 
\smallskip\noindent\textit{\textcolor{reasonFrame}{Critique
confirmed the trace; correction preserves the answer with a minor
terminological fix (alternate $\to$ corresponding angles).}}
\end{correctionbox}

\noindent\textbf{Depth 2} \quad
{\small(\textit{critique detects structural flaw; correction rebuilds from valid basis})}
 
\begin{reasonbox}{Reasoning \quad Answer:~\cmarkw~$28^\circ$}
\small
The corrected reasoning from Step 6 onwards is as follows:\\
Step 6: We have established that $\angle BAC = x^\circ$ due to the
properties of parallel lines and corresponding angles. Now, let's
consider the exterior angle theorem which states that the measure of an
exterior angle of a triangle is equal to the sum of the measures of the
two non-adjacent interior angles. In our case, the exterior angle would
be $\angle BCA$ (which is supplementary to $\angle BAC$).\\
Step 7: Calculate $\angle BCA$:
\[ \angle BCA = 180^\circ - \angle BAC = 180^\circ - x^\circ \]
Step 8: According to the exterior angle theorem:
\[ \angle BCA = \angle ABC + \angle BAC \]
Step 9: Substitute the known values:
\[ 180^\circ - x^\circ = 124^\circ + x^\circ \]
Step 10: Solve for $x$:
\[ 180^\circ - 124^\circ = 2x^\circ,\quad
   56^\circ = 2x^\circ,\quad x^\circ = 28^\circ \]
Therefore, the value of $x$ is $28^\circ$.
\textbf{Final answer:} $\boxed{28}$
\end{reasonbox}
 
\begin{critiquebox}{Self-Critique \quad \textbf{[Error detected]}}
\small
There is a logical mistake in the corrected reasoning. In the corrected
reasoning, there is a slight oversight in the interpretation of the
angles around point $B$. The key mistake lies in not considering the
full circle around point $B$, including the right angle at $A$. This
leads to an incorrect calculation of the total angle sum around $B$ as
$360^\circ$ instead of $180^\circ$ (since two adjacent angles are
supplementary). Correctly accounting for all angles around $B$ would
involve recognising that the sum of the angles on one side of the
transversal equals $180^\circ$, leading to the equation
$124^\circ + x + 90^\circ = 180^\circ$. This correction ensures that
the solution aligns with geometric principles accurately.
 
\smallskip\noindent\textit{\textcolor{critiqueFrame}{%
The critique correctly flags the exterior-angle application as
structurally unsound and prompts a full rebuild.}}
\end{critiquebox}
 
\begin{correctionbox}{Self-Correction \quad Corrected answer:~\cmarkw~$28^\circ$ \textbf{[Reasoning rebuilt]}}
\small
The problem states that $\overline{BC}$ is parallel to the segment
through $A$, and $AB = BC$.\\
\textbf{Step 1:} Identify the given information.
$\overline{BC} \parallel \text{segment through } A$; $AB = BC$.\\
\textbf{Step 2:} Recognize that since $AB = BC$, triangle $ABC$ is
isosceles with base $AC$. Therefore $\angle ABC = \angle ACB$.\\
\textbf{Step 3:} The sum of angles in $\triangle ABC$ is $180^\circ$:
\[ \angle BAC + \angle ABC + \angle ACB = 180^\circ \]
Since $\angle ABC = \angle ACB$, let each equal $y$:
\[ 2y + \angle BAC = 180^\circ \]
\textbf{Step 4:} Since $\overline{BC} \parallel \text{segment through }A$,
corresponding angles are equal; therefore $\angle BAC = 124^\circ$.\\
\textbf{Step 5:} Substitute:
\[ 2y + 124^\circ = 180^\circ,\quad 2y = 56^\circ,\quad y = 28^\circ \]
\textbf{Step 6:} Since $\angle ABC = \angle ACB = y = 28^\circ$,
therefore $x = 28^\circ$.
\textbf{Final answer:} $\boxed{28}$
 
\smallskip\noindent\textit{\textcolor{correctionFrame}{%
Self-correction discards the flawed exterior-angle path entirely and
reconstructs the derivation using the isosceles base-angle property ---
the first time a geometrically valid argument appears in this rollout.}}
\end{correctionbox}

\noindent\textbf{Depth 3}
 
\begin{reasonbox}{Reasoning \quad Answer:~\cmarkw~$28^\circ$}
\small
Continuing from Step 3:\\
Given that $AB = BC$ and $\overline{BC} \parallel \text{segment through
} A$, we know that triangles $ABC$ and $CBA$ are congruent due to the
properties of parallel lines and isosceles triangles. This means that
$\angle ABC = \angle ACB$.\\
Since $\angle BAC = 124^\circ$ (as given), and knowing that the sum of
angles in a triangle is $180^\circ$:
\[
\begin{aligned}
  \angle ABC + \angle ACB + \angle BAC &= 180^\circ,\\
  2\angle ABC + 124^\circ &= 180^\circ,\\
  \angle ABC &= 28^\circ .
\end{aligned}
\]
Therefore, $x = \angle ABC = 28^\circ$.
\textbf{Final answer:} $\boxed{28}$
\end{reasonbox}
 
\begin{critiquebox}{Self-Critique}
\small
The solution appears correct. There were no identified errors in the
provided solution trace.
\end{critiquebox}
 
\begin{correctionbox}{Self-Correction \quad Corrected answer:~\cmarkw~$28^\circ$}
\small
(Same derivation confirmed.)
$\triangle ABC$ isosceles ($AB = BC$) $\Rightarrow \angle ABC = \angle ACB$;
$\angle BAC = 124^\circ$;
$2\angle ABC + 124^\circ = 180^\circ \Rightarrow \angle ABC = 28^\circ$.
\textbf{Final answer:} $\boxed{28}$
 
\smallskip\noindent\textit{\textcolor{correctionFrame}{%
Correct reasoning from depth~2 is stable; no further correction needed.}}
\end{correctionbox}

\noindent\textbf{Depth 4}
 
\begin{reasonbox}{Reasoning \quad Answer:~\cmarkw~$28^\circ$}
\small
To solve for $x$ where $\overline{BC} \parallel \text{segment through }A$
and $AB = BC$:\\
\textbf{1.} Identify given information:
$\overline{BC} \parallel \text{segment through }A$; $AB = BC$.\\
\textbf{2.} Draw diagram: $\triangle ABC$ with $AB = BC$ is isosceles
with base $AC$.\\
\textbf{3.} Analyse angles: let $\angle ABC = 124^\circ$.\\
\textbf{4.} In an isosceles triangle, $\angle BAC = \angle BCA$.
The sum of angles:
\[
\begin{aligned}
  \angle BAC + \angle BCA + \angle ABC &= 180^\circ,\\
  2\angle BAC + 124^\circ &= 180^\circ,\\
  \angle BAC &= 28^\circ .
\end{aligned}
\]
\textbf{5.} Therefore $x = 28^\circ$.
\textbf{Final answer:} $\boxed{28}$
\end{reasonbox}
 
\begin{critiquebox}{Self-Critique}
\small
The solution appears correct. There are no apparent errors in the
provided solution trace. The extracted answer matches the expected
result based on the diagram and the given conditions.
\end{critiquebox}
 
\begin{correctionbox}{Self-Correction \quad \textbf{Final answer:~\cmarkw~$28^\circ$}}
\small
To solve for $x$: $AB = BC \Rightarrow \triangle ABC$ isosceles
$\Rightarrow \angle BAC = \angle BCA$;
$2\angle BAC + 124^\circ = 180^\circ \Rightarrow \angle BAC = 28^\circ$;
therefore $x = 28^\circ$.
\textbf{Final answer:} $\boxed{28}$
 
\smallskip\noindent\textit{\textcolor{correctionFrame}{%
Rollout converges. The self-critique at depth~2 was the pivotal
intervention: it detected a structural flaw and triggered a correction
that produced a valid derivation, which remained stable for the
remaining two depths.}}
\end{correctionbox}

\begin{summarybox}{Per-Rollout Final Answer Summary \quad $N = 8$,\ depth $= 4$}
\footnotesize
\renewcommand{\arraystretch}{1.25}
\begin{tabular}{@{}clp{4.2cm}@{}}
\toprule
\textbf{R\#} & \textbf{Answer} & \textbf{Outcome / Note} \\
\midrule
0 & $112^\circ$ & \xmark~Wrong. Correct at depths 2--3; critique misses regression at depth~4. \\
1 & $28^\circ$  & \cmark~Correct. Stable. \\
\textbf{2} & $\mathbf{28^\circ}$ & \cmark~\textbf{Correct. Critique triggers correction at depth~2; stable thereafter.} \\
3 & $28^\circ$  & \cmark~Correct. Stable. \\
4 & $28^\circ$  & \cmark~Correct. Stable. \\
5 & $28^\circ$  & \cmark~Correct. Stable. \\
6 & $62^\circ$  & \xmark~Wrong. Incorrect reasoning. \\
7 & $152^\circ$ & \xmark~Wrong. Persistent incorrect reasoning. \\
\midrule
\multicolumn{3}{@{}l}{\textbf{Majority vote:}~$28^\circ$\ (5/8)~\cmark}\\
\bottomrule
\end{tabular}
\end{summarybox}

\end{document}